\documentclass{article}

\RequirePackage[l2tabu, orthodox]{nag}
\usepackage[toc,page]{}
\usepackage{warning}
\usepackage{microtype}

\usepackage{graphicx} 
\usepackage{subcaption} 
\usepackage{algorithm}
\usepackage{algorithmic}
\usepackage[draft]{hyperref}

\usepackage{soul}
\usepackage[normalem]{ulem}

\usepackage[symbol]{footmisc}

\usepackage[accepted]{icml2018}

\usepackage{amsmath}
\usepackage{amsthm}
\usepackage{amsfonts}
\newtheorem{prop}{Proposition}

\usepackage{mathrsfs}
\usepackage{flushend}
\usepackage{booktabs}
\usepackage{lipsum}

\input{def1.set}

\newcommand{\x}{\mathbf{x}}

\newcommand{\y}{\mathbf{y}}

\newcommand{\A}{\mathbf{A}}

\newcommand{\Al}{\mathbf{A}}

\newcommand{\W}{\mathbf{W}}
\newcommand{\Wi}{\W^i}

\newcommand{\T}{\mathbf{T}}
\newcommand{\M}{\mathbf{M}}

\newcommand{\logit}{\bzeta}
\newcommand{\logitj}{\zeta_j}

\newcommand{\temp}{\mathbf{c}}
\newcommand{\m}{\mathbf{m}}

\newcommand{\loss}{\mathcal{L}_{\calT}}

\newcommand{\U}{\mathbf{U}}

\newcommand{\mask}{\M}

\newcommand{\maskT}{\mask_{\T}}
\newcommand{\Rn}{\mathbb{R}^n}

\icmltitlerunning{Gradient-Based Meta-Learning with Learned Layerwise Metric and Subspace}

\begin{document}

\twocolumn[
\icmltitle{Gradient-Based Meta-Learning with Learned Layerwise Metric and Subspace}



\icmlsetsymbol{equal}{*}

\begin{icmlauthorlist}
\icmlauthor{Yoonho Lee}{postech}
\icmlauthor{Seungjin Choi}{postech}
\end{icmlauthorlist}

\icmlaffiliation{postech}{Department of Computer Science and Engineering, Pohang University of Science and Technology, Korea}
\icmlcorrespondingauthor{Yoonho Lee}{einet89@postech.ac.kr}
\icmlcorrespondingauthor{Seungjin Choi}{seungjin@postech.ac.kr}

\icmlkeywords{Machine Learning, ICML}

\vskip 0.3in
]

\printAffiliationsAndNotice{}  

\begin{figure*} 
\begin{subfigure}{.9\columnwidth}
  \centering
  \includegraphics[height=1.6in]{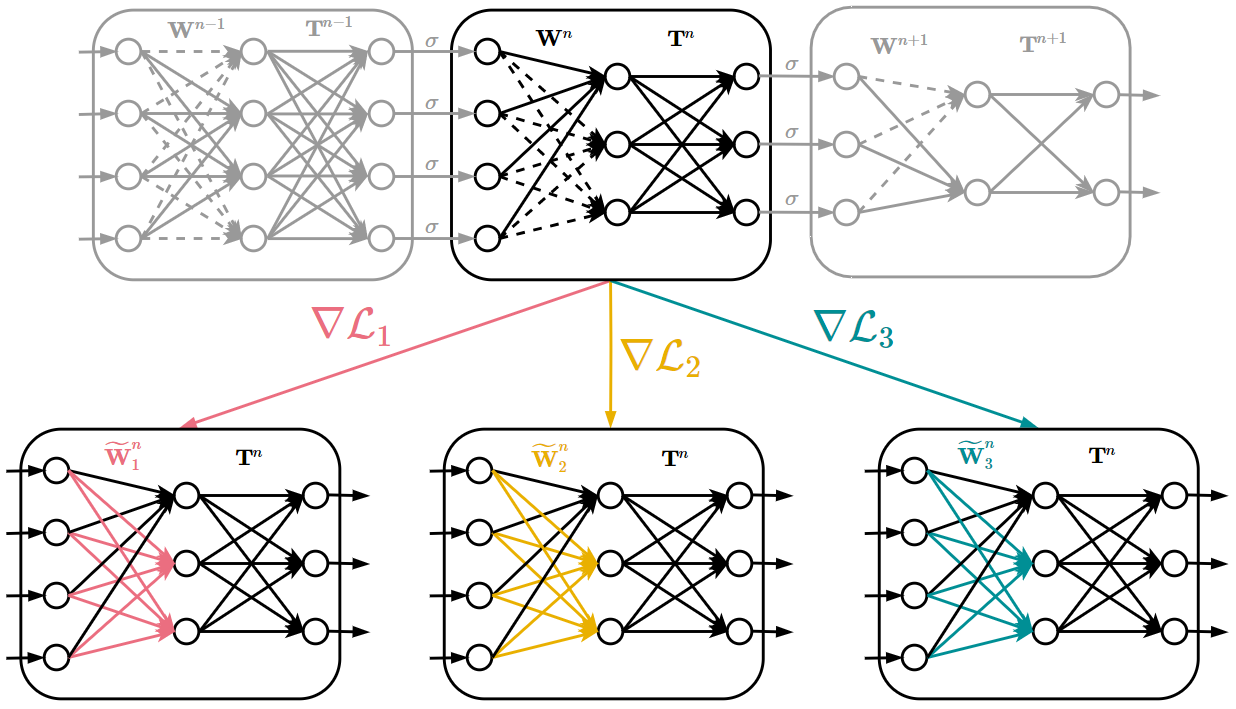}
  \caption{}
  \label{fig:sfig1}
\end{subfigure}%
\begin{subfigure}{.65\columnwidth}
  \centering
  \includegraphics[height=1.6in]{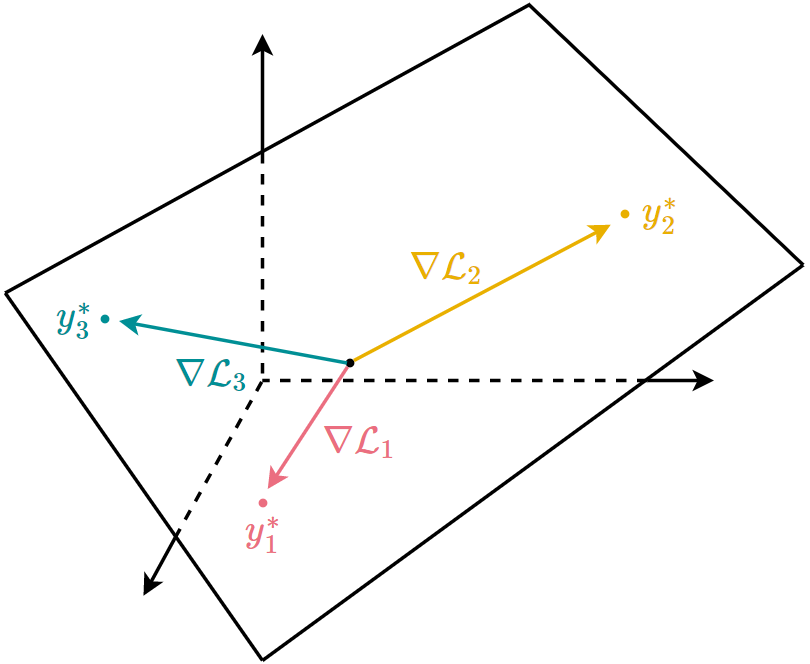}
  \caption{}
  \label{fig:sfig2}
\end{subfigure}
\begin{subfigure}{.45\columnwidth}
  \centering
  \includegraphics[height=1.6in]{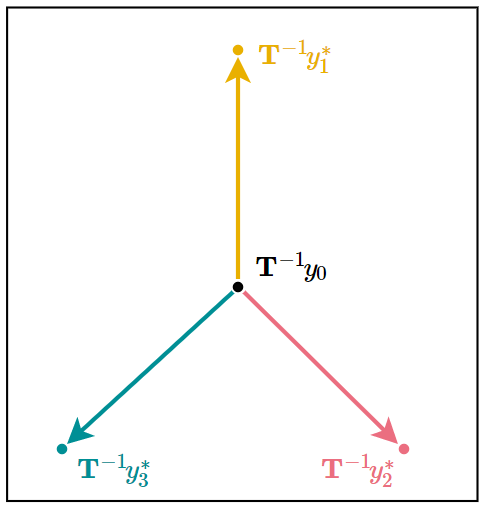}
  \caption{}
  \label{fig:sfig3}
\end{subfigure}
\caption{
Task-specific learning in an MT-net.
(a) A cell (rounded rectangle) consists of two layers.
In addition to initial weights (black), the meta-learner specifies weights to be changed (dotted lines) by task-specific learners (colored).
(b) Activation of this cell has $3$ dimensions, but activation of task-specific learners only change within a subspace (white plane).
(c) The value of $\T$ affects task-specific learning so that gradients of $\W$ are sensitive to task identity.
Best seen in color.
}
\label{fig:fig}
\end{figure*}

\begin{abstract}
Gradient-based meta-learning methods leverage gradient descent to learn the commonalities among various tasks.
While previous such methods have been successful in meta-learning tasks, they resort to simple gradient descent during meta-testing.
Our primary contribution is the {\em MT-net}, which enables the meta-learner to learn on each layer's activation space a subspace that the task-specific learner performs gradient descent on.
Additionally, a task-specific learner of an {\em MT-net} performs gradient descent with respect to a meta-learned distance metric,
which warps the activation space to be more sensitive to task identity.
We demonstrate that the dimension of this learned subspace reflects the complexity of the task-specific learner's adaptation task, and also that our model is less sensitive to the choice of initial learning rates than previous gradient-based meta-learning methods.
Our method achieves state-of-the-art or comparable performance on few-shot classification and regression tasks.
\end{abstract}

\section{Introduction}
\label{sec:introduction}

While recent deep learning methods achieve superhuman performance on various tasks including image classification 
\cite{Krizhevsky2012nips} or playing games \cite{MnihV2015nature}, they can only do so using copious amounts of data and computational resources.
In many problems of interest, learners may not have such luxuries.
\textit{Meta-learning} \cite{SchmidhuberJ87phd,SchmidhuberJ97mlj,ThrunS98book} methods are a potential solution to this problem; 
these methods leverage information gathered from prior learning experience to learn more effectively in novel tasks.
This line of research typically casts learning as a two-level process, each with a different scope.
The \emph{meta-learner} operates on the level of tasks, gathering information from several instances of task-specific learners.
A \emph{task-specific learner}, on the other hand, operates on the level of datapoints, and incorporates the meta-learner's knowledge in its learning process.

Model-agnostic meta-learning (MAML) \cite{FinnC2017icml} is a meta-learning method that directly optimizes the gradient descent procedure of task-specific learners.
All task-specific learners of MAML share initial parameters, and a meta-learner optimizes these initial parameters such that gradient descent 
starting from such initial parameters quickly yields good performance.
An implicit assumption in having the meta-learner operate in the same space as task-specific learners is that the two different scopes of learning
require equal degrees of freedom.

Our primary contribution is the MT-net (Figure 1), a neural network architecture and task-specific learning procedure.
An MT-net differs from previous gradient-based meta-learning methods in that the meta-learner determines a subspace and a corresponding metric 
that task-specific learners can learn in, thus setting the degrees of freedom of task-specific learners to an appropriate amount.
Note that the activation space of the cell shown in Fig.1(b) is $3$-dimensional.
Because the task-specific learners can only change weights that affect two of the three intermediate activations, 
task-specific learning only happens on a subspace with $2$ degrees of freedom.
Additionally, meta-learned parameters $\T$ alter the geometry of the activation space (Fig.1(c)) of task-specific parameters
so that task-specific learners are more sensitive to change in task.

\section{Background}
\label{sec:background}

\subsection{Problem Setup}
\label{subsec:setup}
We briefly explain the meta-learning problem setup which we apply to few-shot tasks.

The problems of $k$-shot regression and classification are as follows.
In the training phase for a meta-learner, we are given a (possibly infinite) set of tasks $\{\calT_1,\calT_2, \calT_3, \ldots\}$.
Each task provides a training set and a test set $\left\{\calD_{\calT_i,train},\calD_{\calT_i,test}\right\}$.
We assume here that the training set $\calD_{\calT_i,train}$ has $k$ examples per class, hence the name $k$-shot learning.
A particular task $\calT \in \{\calT_1,\calT_2, \calT_3, \ldots\}$ is assumed to be drawn from the distribution of tasks $p(\calT)$.
Given a task $\calT \sim p(\calT)$, the task-specific model $f_{\theta_\calT}$ (our work considers a feedforward neural network)
is trained using the dataset $\calD_{\calT,train}$ and 
its corresponding loss $\loss (\theta_\calT,\calD_{\calT,train})$. 
Denote by $\widetilde{\theta}_\calT$ parameters obtained by optimizing $\loss (\theta_\calT,\calD_{\calT,train})$.
Then, the meta-learner $f_{\theta}$ is updated using the feedback from the collection of losses
$\left\{\loss (\widetilde{\theta}_\calT,\calD_{\calT,test}) \right\}_{\calT \sim p(\calT)},$
where the loss of each task is evaluated using the test data $\calD_{\calT,test}$.
Given a new task $\calT_{new}$ (not considered during meta-training),
the meta-learner helps the model $f_{\theta_{\calT_{new}}}$ to quickly adapt to the new task $\calT_{new}$,
by warm-starting the gradient updates.

\subsection{Model-Agnostic Meta-Learning}
We briefly review model-agnostic meta-learning (MAML) \cite{FinnC2017icml}, emphasizing commonalities and differences between MAML and our method.
MAML is a meta-learning method that can be used on any model that learns using gradient descent. 
This method is loosely inspired by fine-tuning, and it learns initial parameters of a network such that the network's loss after a few (usually $1 \sim 5$) gradient steps is minimized.

Consider a model with parameters $\theta$.
MAML alternates between the two updates (\ref{eq:maml_update}) and (\ref{eq:maml_update1})
to determine initial parameters $\theta$ 
for task-specific learners to warm-start the gradient descent updates,
such that new tasks can be solved using a small number of examples.
Each task-specific learner updates its parameters by gradient descent (\ref{eq:maml_update}) 
using the loss evaluated with the training data $\{\calD_{\calT,train}\}$. 
The meta-optimization across tasks (\ref{eq:maml_update1}) is performed such that the parameters $\theta$ are updated
using the loss evaluated with $\{\calD_{\calT,test}\}$.
Note that during meta-optimization (\ref{eq:maml_update1}), the gradient is computed with respect to initial parameters $ \theta $ but the test loss is computed with respect to task-specific parameters $ \widetilde{\theta}_{\calT} $.

\be
\label{eq:maml_update}
\widetilde{\theta}_\calT &\leftarrow& \theta - \alpha \nabla_\theta \loss \left(\theta, \calD_{\calT,train} \right) \\
\theta &\leftarrow& \theta - \beta \nabla_{\theta}
\label{eq:maml_update1}
\left( \sum_{\calT \sim p(\calT)} \loss \left(\widetilde{\theta}_{\calT}, \calD_{\calT,test} \right) \right),
\ee
where $\alpha > 0$ and $\beta > 0$ are learning rates and 
the summation in (\ref{eq:maml_update1}) is computed using minibatches of tasks sampled from $p(\calT)$.

Intuitively, a well-learned initial parameter $\theta$ is close to some local optimum for every task $\calT \sim p(\calT)$.
Furthermore, the update (\ref{eq:maml_update}) is sensitive to task identity in the sense that $\widetilde{\theta}_{\calT_1}$ and $\widetilde{\theta}_{\calT_2}$
have different behaviors for different tasks $\calT_1, \calT_2 \sim p(\calT)$.

Recent work has shown that gradient-based optimization is a universal learning algorithm \cite{FinnC2017arxiv}, 
meaning that any learning algorithm can be approximated up to arbitrary accuracy using some parameterized model and gradient descent.
Thus, no expressiveness is lost by only considering gradient-based learners as in (\ref{eq:maml_update}). 
Note that since MAML operates using a single fixed model, one may have to go through trial and error to find such a good model.

Our method is similar to MAML in that our method also differentiates through gradient update steps to optimize performance after fine-tuning.
However, while MAML assumes a fixed model, our method actually chooses a subset of its weights to fine-tune.
In other words, it (meta-)learns which model is most suitable for the task at hand.
Furthermore, whereas MAML learns with standard gradient descent, a subset of our method's parameters effectively 'warp' the parameter space of the parameters to be learned during meta-testing to enable faster learning.

\begin{figure}[ht!]
\centering\includegraphics[width=\columnwidth]{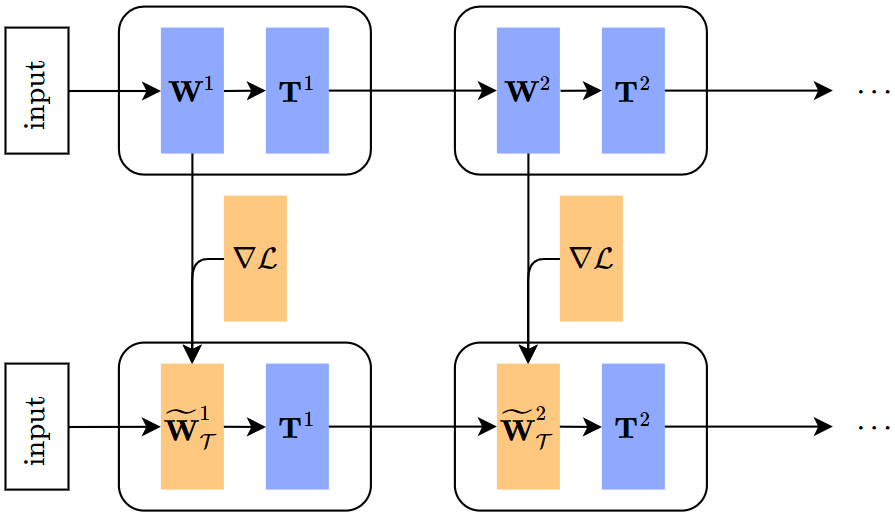}
\caption{A diagram of the adaptation process of a Transformation Network (T-net).
Blue values are meta-learned and shared across all tasks. 
Orange values are different for each task.}
\end{figure}

\section{Meta-Learning Models}
\label{sec:main}

We present our two models in this section: Transformation Networks (T-net) and Mask Transformation Networks (MT-net),
 both of which are trained with gradient-based meta-learning.
A T-net learns a metric in its activation space;
this metric informs each task-specific learner's update direction and step size.
An MT-net additionally learns which subset of its weights to update for task-specific learning.
Therefore, an MT-net learns to automatically assign one of two roles (task-specific or task-mutual) to each of its weights.

\subsection{T-net}
\label{subsec:tnet}

We consider a model $f_{\theta} (\cdot)$ with paramaters $\theta$.
This model consists of $L$ cells, where each cell is parameterized\footnote{
For convolutional cells, $\W$ is a convolutional layer with some size and stride and
and $\T$ is a $1 \times 1$ convolution that doesn't change the number of channels
} as $\T \W$:
\be
\label{eq:tmodel}
\lefteqn{ f_{\theta}(\x)} \nonumber \\
& =  \T^L \W^L \left( \sigma \left( \T^{L-1} \W^{L-1} \left( \ldots 
\sigma \left(  \T^1 \W^1 \x \right) \right) \right) \right),  
\ee
where $\bx \in \Real^{D}$ is an input, and $\sigma(\cdot)$ is a nonlinear activation function.
T-nets get their name from transformation matrices ($\T$) because the linear transformation defined by a $\T$ plays a crucial role in meta-learning.
Note that a cell has the same expressive power as a linear layer.
Model parameters $\theta$ are therefore a collection of $\W$'s and $\T$'s, i.e.,
\bee
\theta =\left\{ \underbrace{\W^1, \ldots, \W^L}_{\theta_{\W}}, 
\underbrace{\T^1, \ldots, \T^L}_{\theta_{\T}}  \right\}.
\eee
Transformation parameters $\theta_{\T}$, which are shared across task-specific models, are determined by the meta-learner.
All task-specific learners share the same initial $\theta_{\W}$ but update to different values since each uses their corresponding train set $\calD_{\calT,train}$.
Thus we denote such (adjusted) parameters for task $\calT$ as $\widetilde{\theta}_{\W,\calT}$.
Though they may look similar, $\calT$ denotes a task while $\T$ denotes a transformation matrix.

\begin{algorithm}[t]
\caption{Transformation Networks (T-net)}
\label{alg:tnet}
\begin{algorithmic}[1]
\REQUIRE $p(\mathcal{T})$
\REQUIRE $\alpha$, $\beta$
\STATE randomly initialize $\theta$
\WHILE{not done}
    \STATE Sample batch of tasks $ \mathcal{T}_i \sim p(\mathcal{T})$ 
    \FORALL{$\mathcal{T}_j$}
        \FOR{$i=1, \cdots, L$}
        	\STATE Compute $\widetilde{\W}_{\calT}$ according to (\ref{eq:theta_wt})
        \ENDFOR
        \STATE $\widetilde{\theta}_{\W, \calT_j} = \{ \widetilde{\W}^{1}_{\calT_j}, \cdots \widetilde{\W}^{L}_{\calT_j} \}$
    \ENDFOR
    \STATE $\theta \leftarrow \theta - \beta \nabla_{\theta} \sum_{j} \loss (\widetilde{\theta}_{\W,\calT_j}, \theta_{\T},\calD_{\calT_j,test})$
\ENDWHILE
\end{algorithmic}
\end{algorithm}

Given a task $\calT$, each $\W$ is adjusted with the gradient update
\be
\label{eq:theta_wt}
\widetilde{\W}_{\calT} \leftarrow \W - \alpha \nabla_{\W} 
\loss \left(\theta_{\W}, \theta_{\T},\calD_{\calT,train} \right).
\ee
Again, $\widetilde{\theta}_{\W,\calT}$ is defined as $\{\widetilde{\W}_{\calT}^1, \ldots, \widetilde{\W}_{\calT}^L \}$.
Using the task-specific learner $\widetilde{\theta}_{\W,\calT}$, the meta-learner improves itself with the gradient update
\be
\label{eq:theta}
\theta  \leftarrow \theta - \beta \nabla_{\theta}
 \left( \sum_{\calT \sim p(\calT)} \loss \left(\widetilde{\theta}_{\W,\calT}, \theta_{\T},\calD_{\calT,test} \right) \right).
\ee
$\alpha > 0$ and $\beta > 0$ are learning rate hyperparameters.
We show our full algorithm in Algorithm \ref{alg:tnet}.

To evaluate on a new task $\calT_{\ast}$, we do the following.
We compute task-specific parameters $\widetilde{\theta}_{\W,\calT_{\ast}}$ using (\ref{eq:theta_wt}),
starting from the meta-learned initial value $\theta_{\W}$.
We report the loss of task-specific parameters $\widetilde{\theta}_{\W,\calT_{\ast}}$ on the test set $\calD_{\calT_{\ast},test}$.

We now briefly examine a single cell:
\bee
\y = \T \W \x,
\eee
where $\x$ is the input to the cell and $\y$ its output.
The squared length of a change in output $\Delta \y = \y^*-\y_0$ is calculated as
\be
\| \Delta \y \|^2 = \left( (\Delta \W) \x \right)^{\top} \left( \T^{\top} \T \right) \left( (\Delta \W) \x \right),
\ee
where $\Delta \W$ is similarly defined as $\W^*-\W_0$.
We see here that the magnitude of $\Delta \y$ is determined by the interaction between $(\Delta \W) \x$ and $\T^\top \T$.
Since a task-specific learner performs gradient descent only on $\W$ and not $\T$, the change in $\y$ resulting from (\ref{eq:theta_wt})
is guided by the meta-learned value $\T^\top \T$.
We provide a more precise analysis of this behavior in Section \ref{sec:analysis}.

\subsection{MT-net}
\label{subsec:mtnet}

\begin{figure}[ht!]
\centering\includegraphics[width=\columnwidth]{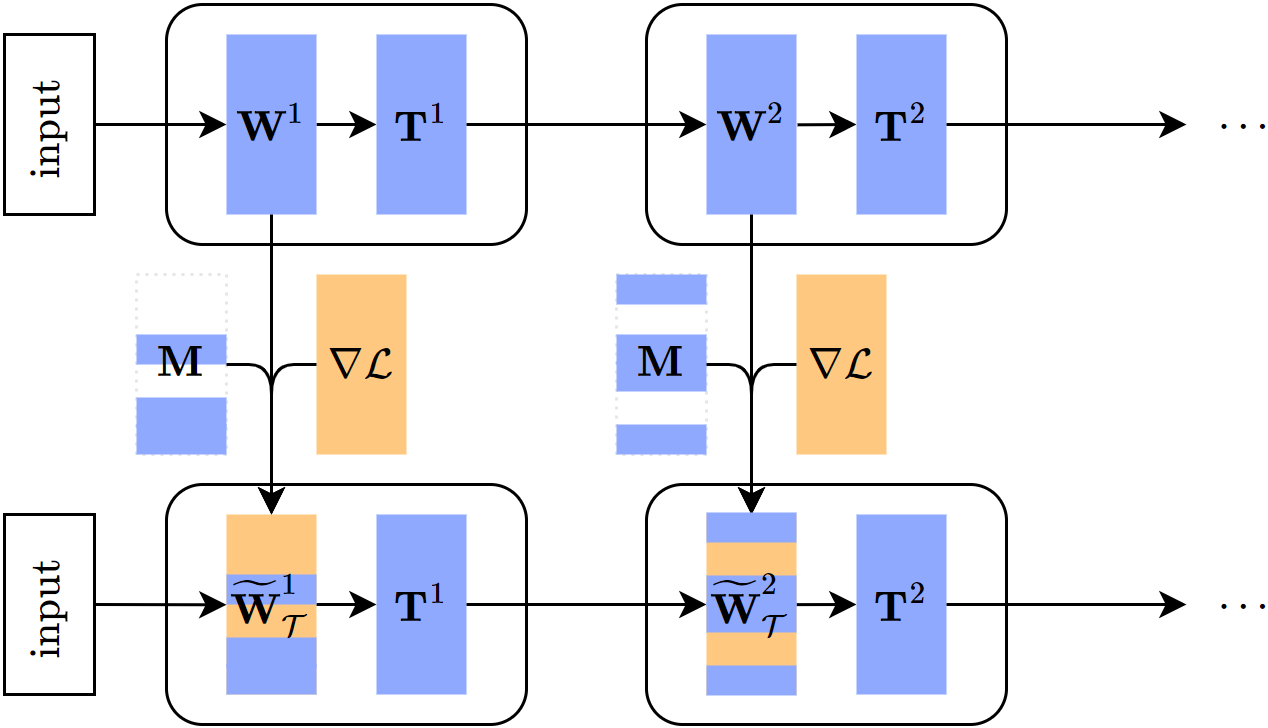}
\caption{
A diagram of the adaptation process of a Mask Transformation Network (MT-net).
Blue values are meta-learned and shared across all tasks. 
Orange values are different for each task.
}
\end{figure}

The MT-net is built on the same feedforward model (\ref{eq:tmodel}) as the T-net:
\be
\label{eq:mtmodel}
\lefteqn{ f_{\theta}(\x)} \nonumber \\
& =  \T^L \W^L \left( \sigma \left( \T^{L-1} \W^{L-1} \left( \ldots 
\sigma \left(  \T^1 \W^1 \x \right) \right) \right) \right).  
\ee
The MT-net differs from the T-net in the binary mask applied to the gradient update 
to determine which parameters are to be updated.
The update rule for task-specific parameters $\widetilde{\W}_{\calT}$ is given by
\be
\label{eq:mt_update}
\widetilde{\W}_{\calT} \leftarrow \W-\alpha \mask \odot \nabla_{\W}  \mathcal{L}(\theta_{\W}, \theta_{\T},\calD_{\calT,train}),
\ee
where $\odot$ is the Hadamard (elementwise) product between matrices of the same dimension.
$\mask$ is a binary gradient mask which is sampled each time the task-specific learner encounters a new task.
Each row of $\mask$ is either an all-ones vector $\bone$ or an all-zeros vector $\bzero$.
We parameterize the probability of row $j$ in $\mask$ being $\bone$ with a scalar variable $\logitj$:
\be
\mask & = & [\m_1, \ldots, \m_n]^\top, \nonumber \\
\label{eq:bern}
\m_j^{\top}  & \sim & \text{Bern} \left(\frac{\exp \left( \logitj \right)}{\exp \left( \logitj \right)+1} \right) \bone^\top,
\ee
where $\text{Bern} (\cdot)$ denotes the Bernoulli distribution.
Each logit $\logit$ acts on a row of a weight matrix $\W$, so weights that contribute to the same immediate activation are updated or not updated together.

We approximately differentiate through the Bernoulli sampling of masks using the Gumbel-Softmax estimator \cite{Jang2017iclr,Maddison2017iclr}:
\be
g_1, g_2 & \sim & \text{Gumbel}(0, 1), \\
\label{eq:gs}
\m_{j}^{\top} & \leftarrow & \frac{\exp \left(\frac{\logitj + g_1}{\temp} \right)}
{\exp \left(\frac{\logitj + g_1}{\temp}\right)+\exp \left(\frac{g_2}{\temp} \right)} \bone^\top,
\ee
where $\temp$ is a temperature hyperparameter.
This reparameterization allows us to directly backpropagate through the mask.
At the limit of $\temp \rightarrow 0$, (\ref{eq:gs}) follows the behavior of (\ref{eq:bern}).

\begin{algorithm}[t]
\caption{Mask Transformation Networks (MT-net)}
\label{alg:mtnet}
\begin{algorithmic}[1]
\REQUIRE $p(\mathcal{T})$
\REQUIRE $\alpha$, $\beta$
\STATE randomly initialize $\theta$
\WHILE{not done}
    \STATE Sample batch of tasks $ \mathcal{T}_i \sim p(\mathcal{T})$ 
    \FORALL{$\mathcal{T}_j$}
        \FOR{$i=1, \cdots, L$}
        	\STATE Sample binary mask $\mask^i$ according to (\ref{eq:gs})
            \STATE Compute $\widetilde{\W}^{i}_{\calT_j}$ according to (\ref{eq:mt_update})

        \ENDFOR
        \STATE $\widetilde{\theta}_{\W, \calT_j} = \{ \widetilde{\W}^{1}_{\calT_j}, \cdots \widetilde{\W}^{L}_{\calT_j} \}$
    \ENDFOR
    \STATE $\theta \leftarrow \theta - \beta \nabla_{\theta} \sum_{j} \loss \left(\widetilde{\theta}_{\W,\calT}, \theta_{\T},\theta_{\logit},\calD_{\calT,test} \right)$
\ENDWHILE
\end{algorithmic}
\end{algorithm}

As in T-nets, we denote the collection of altered weights as $\widetilde{\theta}_{\W,\calT} = \{ \widetilde{\W}^1_{\calT}, \ldots, \widetilde{\W}^L_{\calT} \}$.
The meta-learner learns all parameters $\theta$:
\be
\theta =\left\{ 
\underbrace{\W^1, \ldots, \W^L}_{\theta_{\W}}, 
\underbrace{\T^1, \ldots, \T^L}_{\theta_{\T}},
\underbrace{\logit^1, \ldots, \logit^L}_{\theta_{\logit}},
  \right\}.
\ee
As in a T-net, the meta-learner performs stochastic gradient descent on $\loss \left(\widetilde{\theta}_{\W,\calT}, \theta_{\T},\theta_{\logit},\calD_{\calT,test} \right)$:
\be
\label{eq:theta}
\theta  \leftarrow \theta - \beta \nabla_{\theta}
 \left( \sum_{\calT \sim p(\calT)} \loss \left(\widetilde{\theta}_{\W,\calT}, \theta_{\T},\theta_{\logit},\calD_{\calT,test} \right) \right).
\ee
The full algorithm is shown in Algorithm \ref{alg:mtnet}.

We emphasize that the binary mask used for task-specific learning ($\mask$) depends on meta-learned parameter weights ($\logit$).
Since the meta-learner optimizes the loss in a task after a gradient step ($\ref{eq:mt_update}$), the matrix $\mask$ gets assigned a high probability of
having value $1$ for weights that are meant to encode task-specific information.
Furthermore, since we update $\mask$ along with model parameters $\W$ and $\T$, the meta-learner is incentivized to
learn configurations of $\W$ and $\T$ in which there exists a clear divide between task-specific and task-mutual neurons.

\section{Analysis}
\label{sec:analysis}

In this section, we provide further analysis of the update schemes of T-nets and MT-nets.

We analyse how the activation space of a single cell of a T-net or MT-net behaves during task-specific learning.
More specifically, we make precise how $\W$ encodes a learned curvature matrix.
By using such an analysis to reason about a whole network consisting of several cells, 
we are impliticly approximating the full curvature matrix of the network by
a block-diagonal curvature matrix.
In this approximation, second-order interactions only occur among weights in the same layer (or cell).
Previous works \cite{Heskes2000neuralcomp,Martens2015icml,Desjardins2015nips} have used such an approximation of the curvature of a neural network.

\subsection{T-nets Learn a Metric in Activation Space}
\label{subsec:tnet_analysis}

We consider a cell in a T-net where the pre-activation value $\y$ is given by
\be
\y = \T \W \x = \A \x,
\ee
where $\A=\T \W$ and $\x$ is the input to the cell. 
We omit superscripts throughout this section.

A standard feedforward network resorts to the gradient of a loss function $\loss$ 
(which involves a particular task $\calT \sim p(\calT)$) with respect to the parameter matrix $\A$,
to update model parameters.
In such a case, a single gradient step yields
\be
\y^\text{new} &=& (\A - \alpha \nabla_{\A} \loss) \x  \nonumber \\
&=& \y - \alpha \nabla_{\A} \loss \x.
\ee
The update of a T-net ($\ref{eq:theta_wt}$) results in the following new value of $\y$:
\be
\label{eq:tnet_update}
\y^\text{new} &=& \T \left(\T^{-1} \A - \alpha \nabla_{\T^{-1} \A} \loss \right) \x  \nonumber \\
&=& \y - \alpha  \left( \T \T^{\top} \right) \nabla_{\A} \loss \x,
\ee
where $\T$ is determined by the meta-learner.
Thus, in a T-net, the incremental change of $\y$ is proportional to the negative of the gradient 
$\left( \T  \T^{\top}\right) \nabla_{\A} \loss$, while the standard feedforward net resorts to a step proportional to 
the negative of $ \nabla_{\A} \loss$.
Task-specific learning in the T-net is guided by a full rank metric in each cell's activation space,
which is determined by each cell's transformation matrix $\T$.
This metric $(\T \T^{\top})^{-1}$ warps (scaling, rotation, etc.) the activation space of the model so that in this warped space,
a single gradient step with respect to the loss of a new task yields parameters that are well suited for that task.


\subsection{MT-nets Learn a Subspace with a Metric}
\label{subsec:mtnet_analysis}
We now consider MT-nets and analyze what their update ($\ref{eq:mt_update}$) means from the viewpoint of $\y = \T \W \x = \A \x$.

MT-nets can restrict its task-specific learner to any subspace of its gradient space:

\begin{prop}
\label{prop}
Fix $\x$ and $\A$. 
Let $\y = \T \W \x$ be a cell in an MT-net and let $\logit$ be its corresponding mask parameters.
Let $\U$ be a d-dimensional subspace of $\Rn$ ($d \leq n$).
There exist configurations of $\T, \W,$ and $\logit$ such that the span of $\y^\text{new}-\y$ is $\U$ while satisfying $\A = \T \W$.
\end{prop}

\begin{proof}
See Appendix B.
\end{proof}

This proposition states that $\W, \T$, and $\logit$ have sufficient expressive power to restrict updates of $\y$ to any subspace.
Note that this construction is only possible because of the transformation $\T$; if we only had binary masks $\mask$, we would only be able to restrict gradients to axis-aligned subspaces.

In addition to learning a subspace that we project gradients onto ($\U$), we are also learning a metric in this subspace.
We first provide an intuitive exposition of this idea.

We unroll the update of an MT-net as we did with T-nets in (\ref{eq:tnet_update}):
\begin{align}
\y^\text{new} =& \T ((\T^{-1} \A - \alpha \mask \odot \nabla_{\T^{-1} \A} \loss) \x) \nonumber \\
=& \y - \alpha \T (\mask \odot (\T^{\top} \nabla_{\A} \loss)) \x \nonumber \\
=& \y - \alpha \T (\maskT \odot \T^{\top}) \nabla_{\A} \loss \x \nonumber \\
=& \y - \alpha (\T \odot \maskT^{\top}) (\maskT \odot \T^{\top}) \nabla_{\A} \loss \x.
\label{sumnet_metric}
\end{align}

Where $\maskT$ is an $m \times m$ matrix which has the same columns as $\mask$.
Let's denote $\T_M=\maskT \odot \T^{\top}$.
We see that the update of a task-specific learner in an MT-net performs the update $\T_M^\top \T_M \nabla_{\A} \loss$.
Note that $\T_M^\top \T_M$ is an $n \times n$ matrix that only has nonzero elements in rows and columns where $\m$ is $1$.
By setting appropriate $\logit$, we can view $\T_M^\top \T_M$ as a full-rank $d \times d$ metric tensor.

This observation can be formally stated as:
\begin{prop}
\label{prop2}
Fix $\x$, $\A$, and a loss function $\loss$.
Let $\y = \T \W \x$ be a cell in an MT-net and let $\logit$ be its corresponding mask parameters.
Let $\U$ be a d-dimensional subspace of $\Rn$, and $g(\cdot , \cdot)$ a metric tensor on $\U$.
There exist configurations of $\T, \W,$ and $\logit$ such that the vector $\y^\text{new}-\y$ is in 
the steepest direction of descent on $\loss$ with respect to the metric $g(\cdot , \cdot)$.
\end{prop}

\begin{proof}
See Appendix B.
\end{proof}
Therefore, not only can MT-nets project gradients of task-specific learners onto a subspace of the pre-activation ($\y$) space,
they can also learn a metric in that subspace and thereby learning a low-dimensional linear embedding of the activation space.
The MT-net update (\ref{eq:mt_update}) is gradient descent in this low-dimensional embedding,
so the meta-objective shown in (\ref{eq:theta}) is minimized when gradient descent in this embedding requires few steps to converge and is sensitive to task identity.

\section{Related Work}

A successful line of research in few-shot learning uses feedforward neural networks as learners.
These approaches learn update rules \cite{RaviS2017iclr, Li2016nips, Andrychowicz2016nips} or directly generate weights \cite{Ha2017iclr}.
A related research direction is to learn initial parameters \cite{FinnC2017icml} while fixing the learning rule to gradient descent,
 or additionally learning learning rates for each weight \cite{Li2017arxiv}.
\cite{Grant2018iclr} interprets such gradient-based meta-learning as hierarchical bayesian inference, and
\cite{FinnC2017arxiv} states that such methods are expressive enough to approximate any learning algorithm.

Our work is closely related to this line of research.
Unlike previous work, MT-nets learn how many degrees of freedom the task-specific learner should have at meta-test time.
Additionally, while MT-nets learn update rules, these update rules are directly embedded in the network itself instead of being stored in a separate model.

Distance metric learning \cite{XingEP2002nips, WeinbergerKQ2005nips} methods learn a distance function between datapoints.
Similarly, MT-nets learn a full metric matrix.
Whereas those methods required constrained optimization techniques to enforce that the learned matrix represents a metric,
our parameterization allows us to directly learn such a metric using gradient descent.
Recently, neural networks have been used to learn a metric between images\cite{KochG2015icml, VinyalsO2016nips, SnellJ2017nips},
achieving state-of-the-art performance on few-shot classification benchmarks.
Unlike these methods, we learn a metric in feature space instead of input space.
Our method applies to a larger class of problems including regression and reinforcement learning, since all MT-nets require is a differentiable loss function.

Another line of research in few-shot learning is to use a recurrent neural network (RNN) as a learner \cite{SantoroA2016icml, MunkhdalaiT2017icml}.
Here, the meta-learning algorithm is gradient descent on an RNN, and the learning algorithm is the update of hidden cells.
The (meta-learned) weights of the RNN specify a learning strategy, which processes training data and uses the resulting hidden state vector to make decisions about test data.
A recent work that uses temporal convolutions for meta-learning\cite{Mishra2018iclr} is also closely related to this line of research.

\section{Experiments}
We performed experiments to answer:
\begin{itemize}
\setlength\itemsep{0em}
	\item Do our novel components ($\T \W, \mask$ etc) improve meta-learning performance? (6.1)
	\item Is applying a mask $\mask$ row-wise actually better than applying one parameter-wise? (6.1)
	\item To what degree does $\T$ alleviate the need for careful tuning of step size $\alpha$? (6.2)
	\item In MT-nets, does learned subspace dimension reflect the difficulty of tasks? (6.3)
	\item Can T-nets and MT-nets scale to large-scale meta-learning problems? (6.4)
\end{itemize}

Most of our experiments were performed by modifying the code accompanying \cite{FinnC2017icml},
and we follow their experimental protocol and hyperparameters unless specified otherwise.

\subsection{Toy Regression Problem}
\label{subsec:sine}
\begin{table}[t]
\begin{minipage}{\columnwidth}
\label{tab:sine}
  \centering
\begin{tabular}{llll}
\specialrule{.7pt}{1pt}{1pt}
    Models & 5-shot & 10-shot & 20-shot\\
    \midrule
    MAML\footnote{\label{metasgd} Reported by \cite{Li2017arxiv}.} &  1.07 $\pm$ 0.11 & 0.71 $\pm$ 0.07 & 0.50 $\pm$ 0.05\\
    Meta-SGD\footref{metasgd} & 0.88 $\pm$ 0.14 & 0.53 $\pm$ 0.09 & 0.35 $\pm$ 0.06\\
    \midrule
    M-net-full & 0.91 $\pm$ 0.09 & 0.63 $\pm$ 0.07 & 0.38 $\pm$ 0.04\\
    M-net & 0.88 $\pm$ 0.09 & 0.60 $\pm$ 0.06 & 0.41 $\pm$ 0.04\\
    T-net & 0.83 $\pm$ 0.08 & 0.56 $\pm$ 0.06 & 0.38 $\pm$ 0.04\\
    MT-net-full & 0.81 $\pm$ 0.08 & 0.51 $\pm$ 0.05 & 0.35 $\pm$ 0.04\\
    MT-net & \textbf{0.76 $\pm$ 0.09} & \textbf{0.49 $\pm$ 0.05} & \textbf{0.33 $\pm$ 0.04}\\
\specialrule{.7pt}{1pt}{1pt}
  \end{tabular}
  \caption{
  Loss on sine wave regression.
  Networks were meta-trained using 10-shot regression tasks.
  Reported losses were calculated after adaptation using various numbers of examples.
  }
\end{minipage}
\end{table}

\begin{figure*}[t]
\begin{subfigure}{.5\columnwidth}
  \centering
  \includegraphics[height=1.3in]{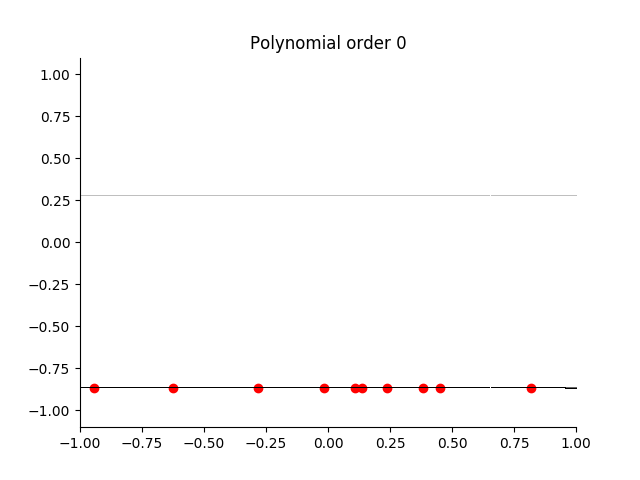}
\end{subfigure}%
\begin{subfigure}{.5\columnwidth}
  \centering
  \includegraphics[height=1.3in]{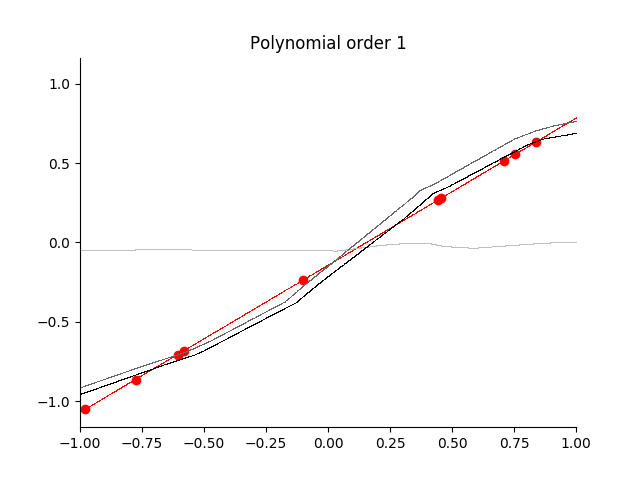}
\end{subfigure}%
\begin{subfigure}{.5\columnwidth}
  \centering
  \includegraphics[height=1.3in]{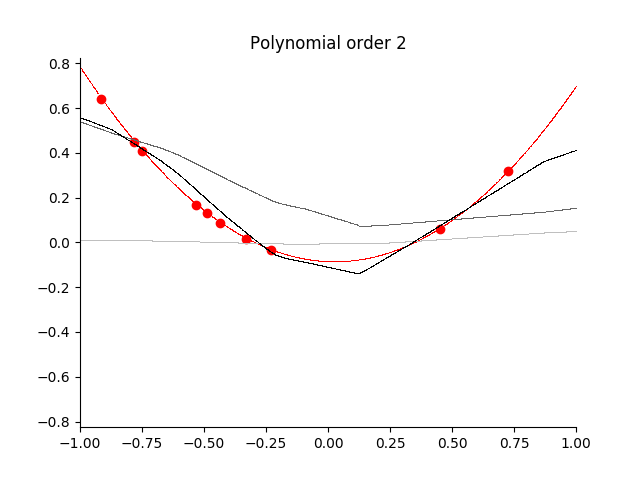}
\end{subfigure}%
\begin{subfigure}{.5\columnwidth}
  \centering
  \includegraphics[height=1.3in]{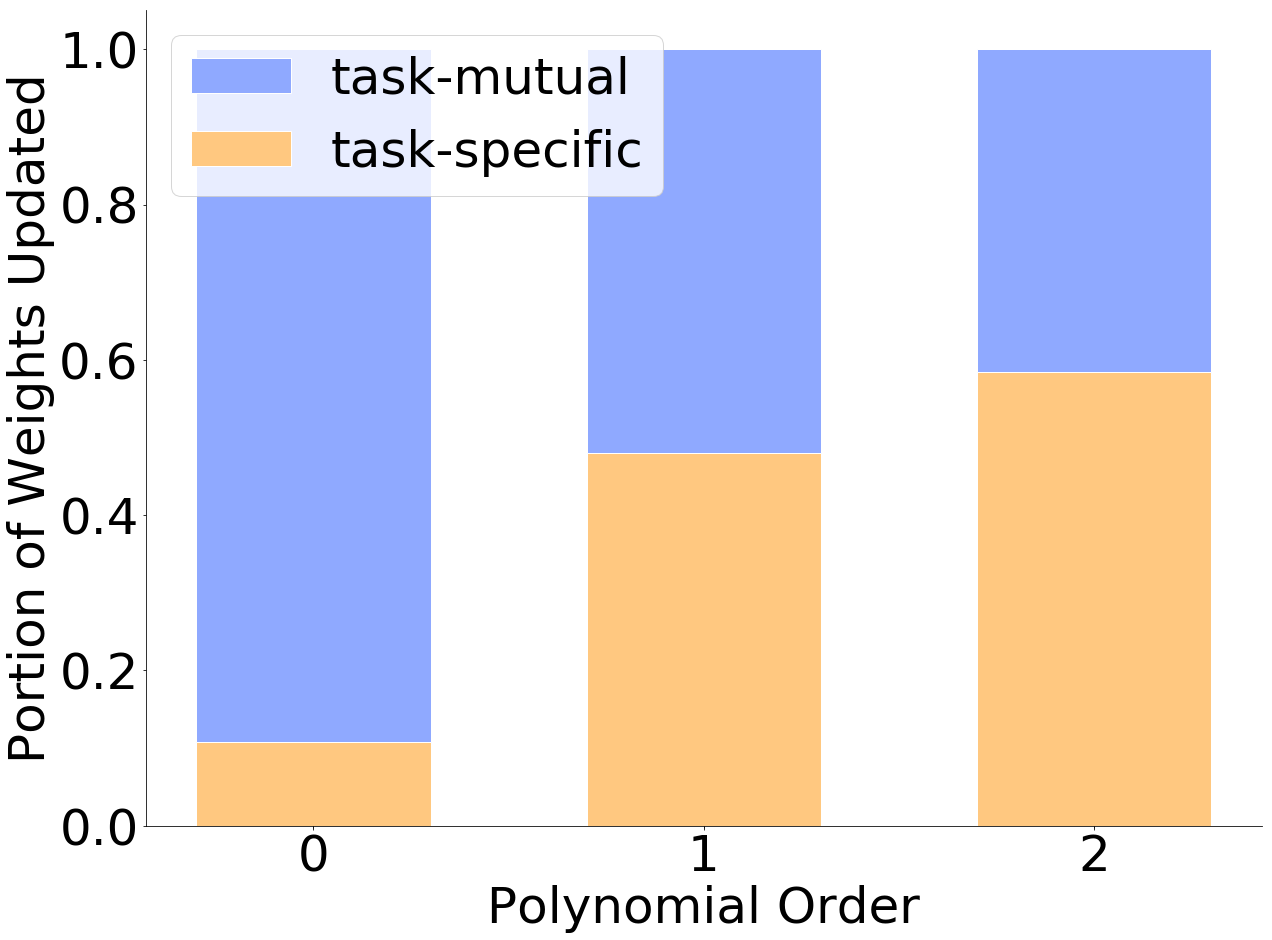}
\end{subfigure}%

\begin{subfigure}{2\columnwidth}
  \includegraphics[width=.75\columnwidth]{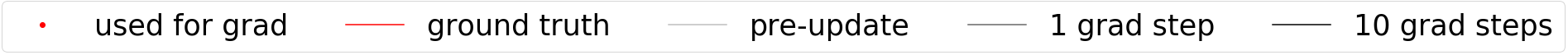}
\end{subfigure}%
\caption{
10-shot regression tasks to sets of polynomials of various degrees.
MT-nets choose to update a larger fraction of weights as the set of tasks gets more complex.
}
\end{figure*}
We start with a $K$-shot regression problem and compare results to previous meta-learning methods \cite{FinnC2017icml, Li2017arxiv}.
The details of our regression task are the same as \cite{Li2017arxiv}.
Each individual task is to regress from the input x to the output y of a sine function
\begin{align}
y(x) = A \sin (wx + b)
\end{align}
For each task, $A,w,b$ are sampled uniformly from $[0.1,5.0], [0.8,1.2],$ and $[0,\pi]$, respectively.
Each task consists of $K \in \{5, 10, 20\}$ training examples and $10$ testing examples.
We sample $x$ uniformly from $[-5.0, 5.0]$ for both train and test sets.
Our regressor architecture has two hidden cells each with activation size $40$.
After every $\T$ is a ReLU nonlinearity.
The loss function is the mean squared error (MSE) between the regressor's prediction $f(x)$ and the true value $y(x)$.
We used Adam \cite{KingmaDP2015iclr} as our meta-optimizer with a learning rate of $\beta=10^{-3}$.
Task-specifc learners used step size $\alpha=10^{-2}$.
We initialize all $\logit$ to $0$, all $\T$ as identity matrices, and all $\W$ as truncated normal matrices with standard deviation $10^{-2}$.
While we trained our meta-learner with $K=10$ examples, we tested using various numbers of examples ($K \in \{5, 10, 20\}$).

We show results in Table 1.
To see if each of our novel components increased meta-learning performance, we also performed the same experiments with variations of MT-nets.
An M-net uses a mask $\mask$ like an MT-net, but each cell consists of a single matrix $\W$ instead of $\T \W$.
A model with "-full" at the end of its name learns a separate mask parameter for each weight of $\W$ instead of sharing a mask 
among weights that contribute to the same activation.
For example, if $\W$ has size $5 \times 10$, the corresponding $\logit$ in an MT-net would be of dimension $5$,
but in MT-net-full, the dimension of $\logit$ would be $50$.
MT-nets outperform MAML, meta-SGD, and all variations of MT-nets.

\subsection{Robustness to learning rate change}
\begin{table}[t]
  \centering
\label{tab:sinelr}
\begin{tabular}{llll}
\specialrule{.7pt}{1pt}{1pt}
	$\alpha$ & MAML & T-net & MT-net \\
    \midrule
	10 & 171.92 $\pm$ 25.04 & \textbf{4.08 $\pm$ 0.30} & 4.18 $\pm$ 0.30 \\
	1 & 5.81 $\pm$ 0.49 & 4.15 $\pm$ 0.30 & \textbf{0.61 $\pm$ 0.07} \\
	0.1 & 1.05 $\pm$ 0.11 & 0.68 $\pm$ 0.06 & \textbf{0.54 $\pm$ 0.05} \\
	0.01 & 0.71 $\pm$ 0.07 & 0.56 $\pm$ 0.06 & \textbf{0.49 $\pm$ 0.05} \\
	0.001 & 0.82 $\pm$ 0.08 & \textbf{0.59 $\pm$ 0.06} & \textbf{0.59 $\pm$ 0.06} \\
	0.0001 & 2.54 $\pm$ 0.19 & \textbf{0.62 $\pm$ 0.06} & 0.72 $\pm$ 0.07\\
\specialrule{.7pt}{1pt}{1pt}
  \end{tabular}
  \caption{
  Loss on 10-shot sine wave regression.
  T-nets and MT-nets are bost robust to change in step size $\alpha$.
  This is due to the meta-learned matrix $\T$ inside each cell, which alters the effective step size.
  }
\end{table}
The transformation $\T$ of our method can change the effective step size $\alpha$.
We performed experiments to see how robust our method is to variations in $\alpha$.
We perform the same sinusoid experiment as in section 6.1, but with various step sizes ($\alpha \in \{10^{-4}, 10^{-3}, 10^{-2}, 10^{-1}, 1, 10\}$).
We evaluate on $K=10$ training examples, and all other settings are identical to the experiments in section 6.1.

We show losses after adaptation of both MAML and MT-nets in Table 2.
We can see that MT-nets are more robust to change in step size.
This indicates that as shown in section \ref{subsec:mtnet_analysis}, the matrix $\T$ is capable of warping the parameter space
to recover from suboptimal step size $\alpha$.

\subsection{Task Complexity and Subspace Dimension}

\begin{table*}[t]
  \centering
\begin{minipage}{\textwidth}
  \centering
  \label{tab:class}
  \begin{tabular}{lll}
\specialrule{.7pt}{1pt}{1pt}
    Models & 5-way 1-shot acc. ($\%$) & 20-way 1-shot acc. ($\%$) \\
    \hline
    \textbf{Matching Networks}\cite{VinyalsO2016nips} & 98.1 & 93.8 \\
    \textbf{Prototypical Networks}\cite{SnellJ2017nips}& 97.4 & 92.0 \\
    \textbf{mAP-SSVM}\cite{Triantafillou2017} & 98.6 & 95.4 \\
    \hline
    \textbf{MAML}\cite{FinnC2017icml} & 98.7 $\pm$ 0.4 & 95.8 $\pm$ 0.3 \\
    \textbf{Meta-SGD}\cite{Li2017arxiv} & \textbf{99.53 $\pm$ 0.26} & 95.93 $\pm$ 0.38\\
    \hline
    \textbf{T-net (ours)} & 99.4 $\pm$ 0.3 & 96.1 $\pm$ 0.3 \\
    \textbf{MT-net (ours)} & \textbf{99.5 $\pm$ 0.3} & \textbf{96.2 $\pm$ 0.4} \\
\specialrule{.7pt}{1pt}{1pt}
  \end{tabular} \\
  \vspace{10pt}
  \centering
  \begin{tabular}{ll}
\specialrule{.7pt}{1pt}{1pt}
    Models & 5-way 1-shot acc. ($\%$) \\
    \hline
    \textbf{Matching Networks}\cite{VinyalsO2016nips}\footnote{\label{metalstm} Reported by \cite{RaviS2017iclr}.} & 43.56 $\pm$ 0.84\\
    \textbf{Prototypical Networks}\cite{SnellJ2017nips}\footnote{ Reported results for 5-way 1-shot.} &  46.61 $\pm$ 0.78 \\
    \textbf{mAP-SSVM}\cite{Triantafillou2017} & 50.32 $\pm$ 0.80 \\
    \hline
    \textbf{Fine-tune baseline}\footref{metalstm} &  28.86 $\pm$ 0.54 \\
    \textbf{Nearest Neighbor baseline}\footref{metalstm}&  41.08 $\pm$ 0.70 \\
    \textbf{meta-learner LSTM}\cite{RaviS2017iclr} &  43.44 $\pm$ 0.77 \\
    \textbf{MAML}\cite{FinnC2017icml} &  48.70 $\pm$ 1.84 \\
    \textbf{L-MAML}\cite{Grant2018iclr} &  49.40 $\pm$ 1.83 \\
    \textbf{Meta-SGD}\cite{Li2017arxiv} &  50.47 $\pm$ 1.87 \\
    \hline
    \textbf{T-net (ours)} & 50.86 $\pm$ 1.82 \\
    \textbf{MT-net (ours)} & \textbf{51.70 $\pm$ 1.84} \\
\specialrule{.7pt}{1pt}{1pt}
\end{tabular}
  \centering
  \caption{Few-shot classification accuracy on (top) held-out Omniglot characters and (bottom) test split of MiniImagenet. $\pm$ represents $95\%$ confidence intervals.}
\end{minipage}
\end{table*}
We performed this experiment to see whether the dimension of the learned subspace of MT-nets reflect the underlying complexity of its given set of tasks.

We consider 10-shot regression tasks in which the target function is a polynomial.
A polynomial regression meta-task consists of polynomials of the same order with various coefficients.
To generate a polynomial of order $n$ ($\sum_{i=0}^{n}c_ix^i$), we uniformly sampled $c_0, \ldots, c_n$ from $[-1, 1]$.
We used the same network architecture and hyperparameters as in Section 6.1 and performed $10$-shot regression for polynomial orders $n \in \{0,1,2\}$.
Since the number of free parameters is proportional to the order of the polynomial, 
we expect higher-order polynomials to require more parameters to adapt to.
The fraction of parameters that task-specific learners change is calculated as the expected value of 
$\frac{e^{\tiny -\logit}}{e^{\tiny -\logit}+1}$ over all logits $\logit$.

We show results in Figure 4, and additional results in Appendix C.
The number of weights that the meta-learner of an MT-net sets to be altered increases as the task gets more complex.
We interpret this as the meta-learner of MT-nets having an effect akin to Occam's razor:
it selects a task-specific model of just enough complexity to learn in a set of tasks.
This behavior emerges even though we do not introduce any additional loss terms to encourage such behavior.
We think this is caused by the noise inherent in stochastic gradient descent.
Since the meta-learner of an MT-net can choose whether or not to perform gradient descent in a particular direction, it is incentivized not to do so in directions that are not model-specific, because doing so would introduce more noise into the network parameters and thus (in expectation) suffer more loss.

\subsection{Classification }

To compare the performance of MT-nets to prior work in meta-learning, 
we evaluate our method on few-shot classification on the Omniglot \cite{Lake2015science} and MiniImagenet \cite{RaviS2017iclr} datasets.
We used the miniImagenet splits proposed by \cite{RaviS2017iclr} in our experiments.

Our CNN model uses the same architecture as \cite{FinnC2017icml}.
The model has $4$ modules: each has $3 \times 3$ convolutions and $64$ filters, followed by batch normalization \cite{Ioffe2015icml}.
As in \cite{FinnC2017icml}, we used 32 filters per layer in miniImagenet.
Convolutions have stride $2 \times 2$ on Omniglot, and $2 \times 2$ max-pooling is used after batch normalization
instead of strided convolutions on MiniImagenet.
We evaluate with 3, 5, and 10 gradient steps for Omniglot 5-way, Omniglot 20-way, and miniImagenet 5-way, respectively.

Results are shown in Table 3.
MT-nets achieve state-of-the-art or comparable performance on both problems.
Several works \cite{Mishra2018iclr,MunkhdalaiT2017icml,Sung2017arxiv} have reported improved performance on MiniImagenet using a significantly more expressive architecture.
We only report methods that have equal or comparable expressiveness to the model first described in \cite{VinyalsO2016nips}.
Not controlling for network expressivity, the highest reported accuracy so far on 5-way 1-shot miniImagenet classification is $57.02$ \cite{Sung2017arxiv}.

\section{Conclusion}

We introduced T-nets and MT-nets.
One can transform any feedforward neural network into an MT-net, so any future architectural advances can take advantage of our method.
Experiments showed that our method alleviates the need for careful tuning of the learning rate in few-shot learning problems 
and that the mask $\mask$ reflects the complexity of the set of tasks it is learning to adapt in.
MT-nets also showed state-of-the-art performance in a challenging few-shot classification benchmark (MiniImagenet).

While we think MT-nets are a gradient-based meta-learning method, our analysis has shown that it has some interesting commonalities with optimizer learning methods such as \cite{RaviS2017iclr}.
We will investigate this connection between two seemingly disparate approaches to meta-learning in future work.

One of the biggest weaknesses of deep networks is that they are very data intensive.
By learning what to learn when a new task is encountered, we can train networks with high capacity using a small amount of data.
We believe that designing effective gradient-based meta-learners will be beneficial not just for the few-shot learning setting, 
but also machine learning problems in general.

\clearpage

\section*{Acknowledgements}
SC was supported by Samsung DS Software Center, Samsung Research, and Naver.

\nocite{SnellJ2017nips}
\nocite{RaviS2017iclr}
\nocite{FinnC2017icml}
\nocite{FinnC2017arxiv}
\nocite{VinyalsO2016nips}
\nocite{AmariS98neco}
\nocite{KochG2015icml}
\nocite{SantoroA2016icml}
\nocite{RobbinsH51ams}
\nocite{ThrunS98book}
\nocite{MnihV2015nature}
\nocite{MunkhdalaiT2017icml}
\nocite{Sung2017arxiv}
\nocite{Edwards2017iclr}
\nocite{Kaiser2017iclr}
\nocite{Garcia2017arxiv}
\nocite{Li2017aaai}
\nocite{tsai2017arxiv}
\nocite{fort2017arxiv}
\nocite{Ren2018iclr}
\nocite{Hariharan2017iccv}
\nocite{Mishra2018iclr}
\nocite{Maddison2017iclr}

\bibliography{sjc}
\bibliographystyle{icml2018}

\clearpage


\end{document}


\begin{appendices}
\section{Further Experimental Details}
We used the same hyperparameters as \cite{FinnC2017icml} and terminated training after the same number of examples specifid in \cite{FinnC2017icml}.
We used a temperature $\temp$ of $1$, which was recommended in \cite{Jang2017iclr}.
We initialized $\T$ to be an identity matrix and all $\logit$ to zero.

Compated to MAML \cite{FinnC2017icml}, training a convolutional MT-net takes roughly $0.4$ times longer (omniglot 40k steps took 7h 19m for MT-net and 5h 14m for MAML).
This gap is fairly small because $1 \times 1$ convolutions require little compute compared to regular convolutions.
This gap is larger (roughly $1.1$ times) for fully connected MT-nets.
We additionally observed that MT-nets take less training steps to converge compared to MAML.

We provide our official implementation of MT-nets at \url{https://github.com/yoonholee/MT-net}.

\bibliographystyle{abbrv}
\bibliography{sjc}

\vspace{1.2cm}

\section{Proofs for Section 4}
\subsection{MT-nets Learn a Subspace}
\setcounter{prop}{0}

\begin{prop}
Fix $\x$ and $\A$. 
Let $\U$ be a d-dimensional subspace of $\Rn$ ($d \leq n$).
There exist configurations of $\T, \W,$ and $\logit$ such that the span of $\y^\text{new}-\y$ is $\U$ while satisfying $\A = \T \W$.
\end{prop}
\begin{proof}
We show by construction that Proposition 1 is true.

Suppose that $\{\bv_1, \bv_2, \ldots, \bv_n\}$ is a basis of $\Real^n$ such that $\{\bv_1, \bv_2, \ldots, \bv_d\}$ is a basis of $\U$.
Let $\T$ be the $n \times n$ matrix $[\bv_1, \bv_2, \ldots, \bv_n]$.
$\T$ is invertible since it consists of linearly independent columns.
Let $\W = \T^{-1} \A$ and let $\logit_1, \logit_2, \ldots, \logit_d \rightarrow \infty$ and $\logit_{d+1}, \ldots, \logit_n \rightarrow -\infty$.
The resulting mask $\M$ that $\logit$ generates is a matrix with only ones in the first $d$ rows and zeroes elsewhere.
\be
\y^\text{new} - \y = \T (\W^\text{new} - \W) \x \nonumber \\
= \T (\mask \odot \nabla_\W \loss) \x
\ee
Since all but the first $d$ rows of $\mask$ are $\mathbf{0}$, $(\mask \odot \nabla_\W \loss) \x$ is an $n$-dimensional vector in which nonzero elements can
only appear in the first $d$ dimensions.
Therefore, the vector $\T (\mask \odot \nabla_\W \loss) \x$ is a linear combination of $\{\bv_1, \bv_2, \ldots, \bv_d\}$.
Thus the span of $\y^\text{new}-\y$ is $\U$.
\end{proof}

\subsection{MT-nets Learn a Metric in their Subspace}
\begin{prop}
Fix $\x$, $\A$, and a loss function $\loss$.
Let $\U$ be a d-dimensional subspace of $\Rn$, and $g(\cdot , \cdot)$ a metric tensor on $\U$.
There exist configurations of $\T, \W,$ and $\logit$ such that the vector $\y^\text{new}-\y$ is in 
the steepest direction of descent on $\loss$ with respect to the metric $du$.
\end{prop}

\begin{proof}
We show Proposition 2 is true by construction as well.

We begin by constructing a representation for the arbitrary metric tensor $g(\cdot , \cdot)$.
Let $\{\bv_1, \bv_2, \ldots, \bv_n\}$ be a basis of $\Rn$ such that $\{\bv_1, \bv_2, \ldots, \bv_d\}$ is a basis of $\U$.
Vectors $\bu_1, \bu_2 \in \U$ can be expressed as $\bu_1 = \sum_{i=0}^d c_{1i} \bv_i$ and $\bu_2 = \sum_{i=0}^d c_{2i} \bv_i$.
We can express any metric tensor $g(\cdot , \cdot)$ using such coefficients $c$:
\be
g(\bu_1, \bu_2 ) = 
\underbrace{
\begin{bmatrix} 
c_{11} & \dots & c_{1d}
\end{bmatrix}
}_{\bc_1^\top}
\underbrace{
\begin{bmatrix} 
g_{11} & \dots & g_{1d}\\
\vdots & \ddots & \vdots \\
g_{d1} & \dots & g_{dd}
\end{bmatrix}
}_{\bG}
\underbrace{
\begin{bmatrix} 
c_{21}\\
\vdots \\
c_{2d}
\end{bmatrix}
}_{\bc_2},
\ee
where $\bG$ is a positive definite matrix.
Because of this, there exists an invertible $d \times d$ matrix $\bH$ such that $\bG=\bH^\top \bH$.
Note that $g(\bu_1, \bu_2 ) = (\bH \bc_1)^\top (\bH \bc_2)$: 
the metric $g(\cdot , \cdot)$ is equal to the inner product after multiplying $\bH$ to given vectors $\bc$.

Using $\bH$, we can alternatively parameterize vectors in $\U$ as
\be
\bu_1 &=&
\underbrace{
\begin{bmatrix} 
\bv_1 & \dots & \bv_d
\end{bmatrix}
}_{\bV}
\bc_1
\\
&=& \bV \bH^{-1} \left( \bH \bc_1 \right).
\ee
Here, we are using $\bH \bc_1$ as a $d$-dimensional parameterization and the 
columns of the $n \times d$ matrix $\bV \bH^{-1}$ as an alternative basis of $\U$.

Let $\bv_1^H, \ldots, \bv_d^H$ be the columns of $\bV \bH^{-1}$, and set $\T=[\bv_1^\bH, \ldots, \bv_d^\bH, \bv_{d+1}, \ldots, \bv_n]$.
Since $\bH$ is invertible, $\{\bv_1^\bH, \ldots, \bv_d^\bH\}$ is a basis of $\U$ and thus $\T$ is an invertible matrix.
As in Proposition 1, set $\W = \T^{-1} \A$, $\logit_1, \logit_2, \ldots, \logit_d \rightarrow \infty$, 
and $\logit_{d+1}, \ldots, \logit_n \rightarrow -\infty$.
Note that this configuration of $\logit$ generates a mask $\mask$ that projects gradients onto the first $d$ rows, 
which will later be multiplied by the vectors $\{\bv_1^\bH, \ldots, \bv_d^\bH\}$.

We can express $\y$ as $\y=V \bc_\y=\bV \bH^{-1} (\bH \bc_\y)$, where $\bc_\y$ is again a $d$-dimensional vector.
Note that $\bV \bH^{-1}$ is constant in the network and change in $\W$ only affects $\bH \bc_\y$.
Since $\nabla_{\W} \loss = (\nabla_{\W \x} \loss) \x^\top$, 
the task-specific update is in the direction of steepest descent of $\loss$ in the space of $\bH \bc_y$ (with the Euclidean metric).
This is exactly the direction of steepest descent of $\loss$ in $\U$ with respect to the metric $g(\cdot , \cdot)$.
\end{proof}

\clearpage
\section{Additional Experiments}
\label{sec:add_poly}
\begin{figure}[htb]
\setlength{\unitlength}{.25\columnwidth}
\begin{picture}(.5,3.9) \linethickness{0.5pt}
\put(0.0,0.0){\includegraphics[width=0.27\columnwidth]{figures/0/0}}
\put(0.0,.8){\includegraphics[width=0.27\columnwidth]{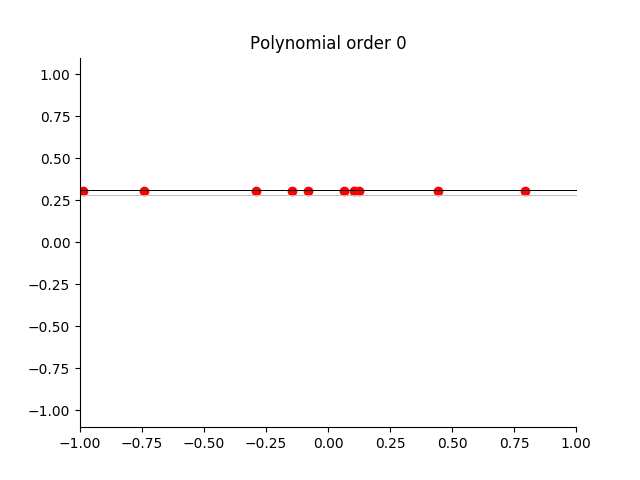}}
\put(0.0,1.6){\includegraphics[width=0.27\columnwidth]{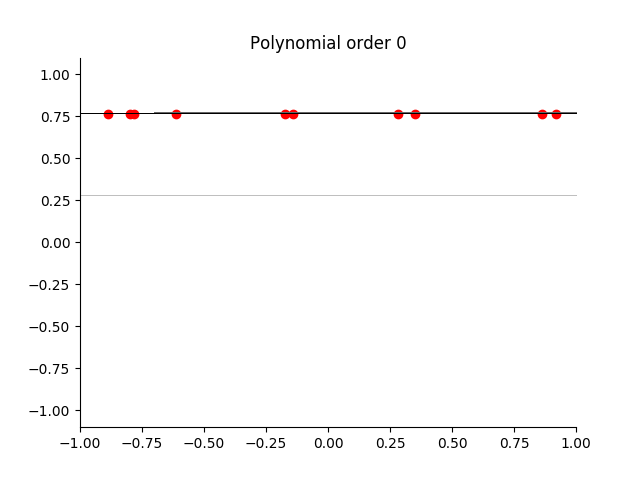}}
\put(0.0,2.4){\includegraphics[width=0.27\columnwidth]{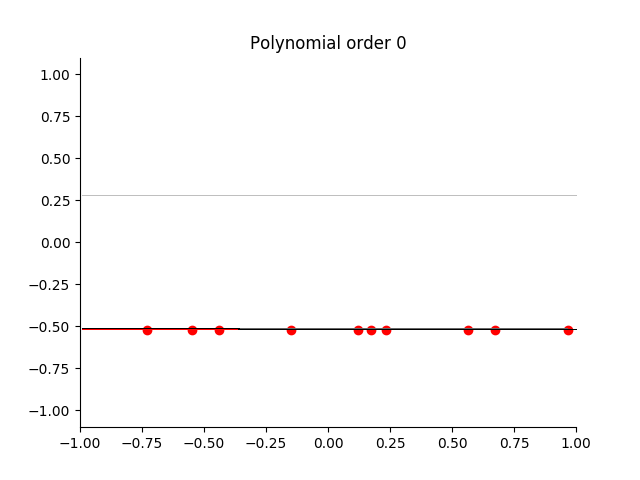}}
\put(0.0,3.2){\includegraphics[width=0.27\columnwidth]{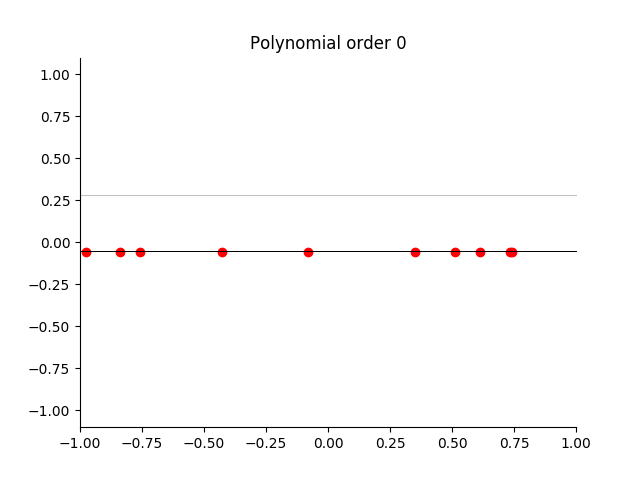}}

\put(1.33,0.0){\includegraphics[width=0.27\columnwidth]{figures/1/0}}
\put(1.33,.8){\includegraphics[width=0.27\columnwidth]{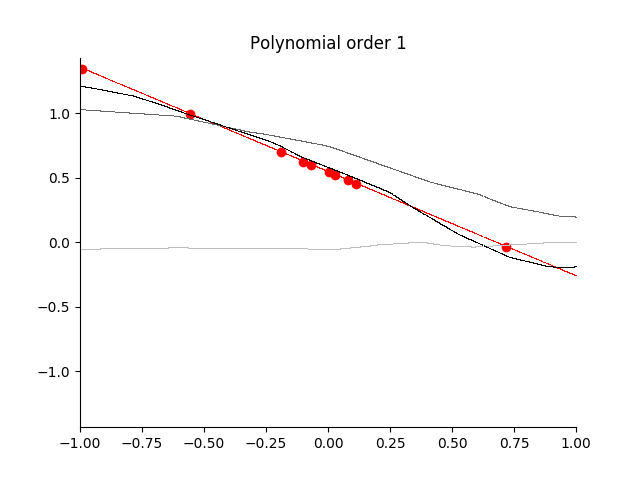}}
\put(1.33,1.6){\includegraphics[width=0.27\columnwidth]{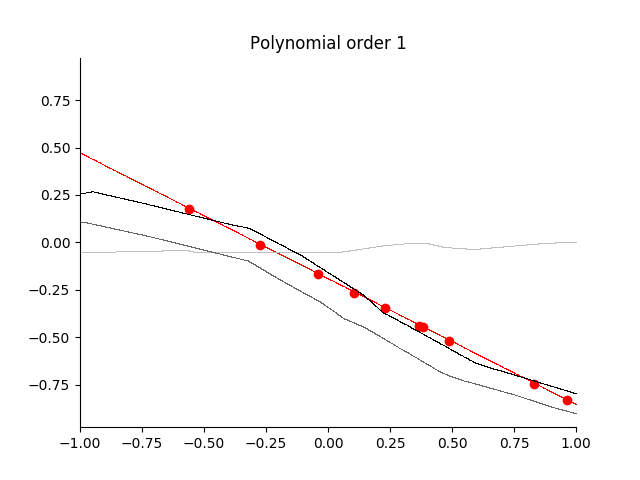}}
\put(1.33,2.4){\includegraphics[width=0.27\columnwidth]{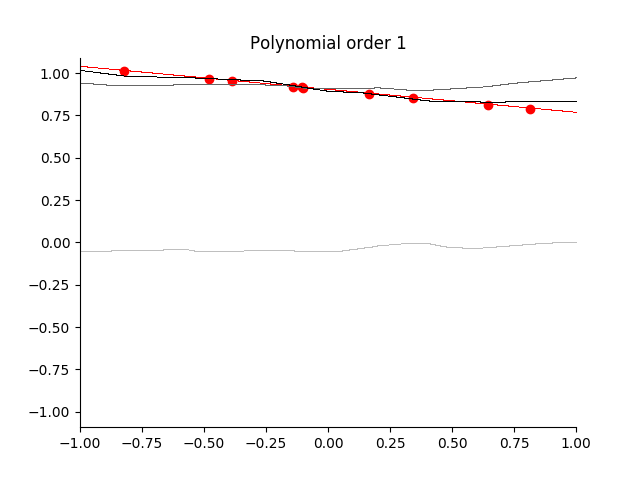}}
\put(1.33,3.2){\includegraphics[width=0.27\columnwidth]{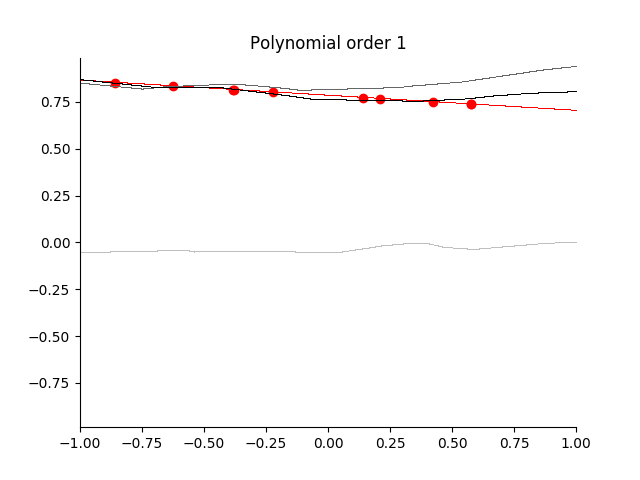}}

\put(2.66,0.0){\includegraphics[width=0.27\columnwidth]{figures/2/0}}
\put(2.66,.8){\includegraphics[width=0.27\columnwidth]{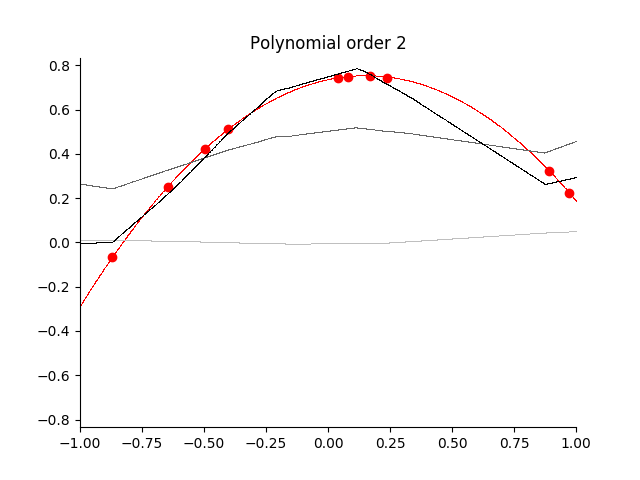}}
\put(2.66,1.6){\includegraphics[width=0.27\columnwidth]{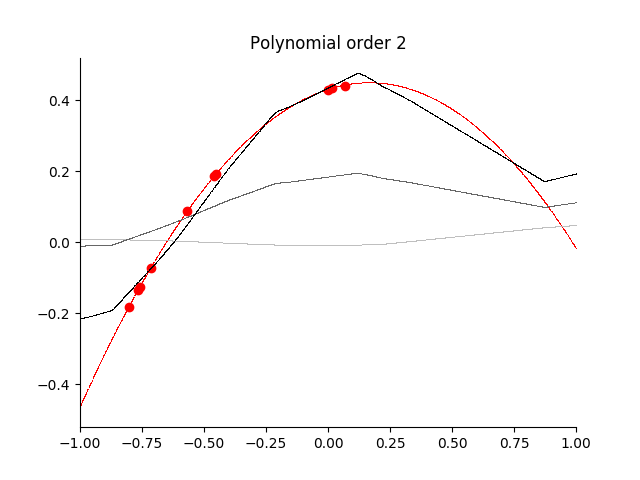}}
\put(2.66,2.4){\includegraphics[width=0.27\columnwidth]{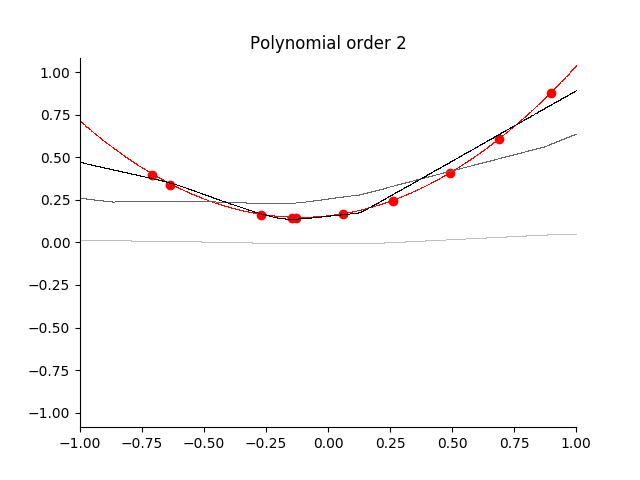}}
\put(2.66,3.2){\includegraphics[width=0.27\columnwidth]{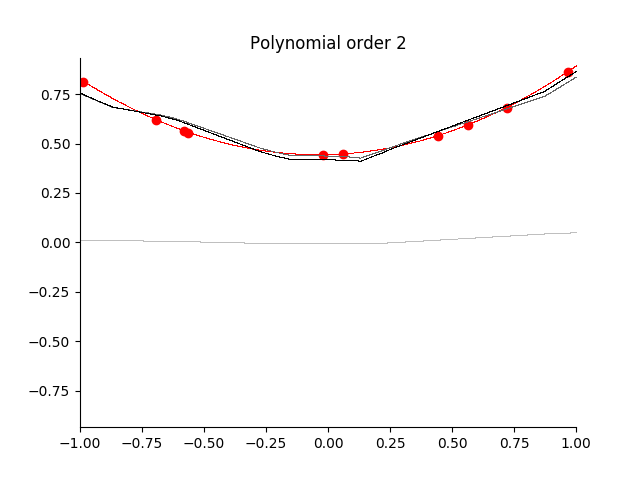}}

\put(0.0,-0.2){\includegraphics[width=.95\columnwidth]{figures/legend}}

\end{picture}
\caption{Additional qualitative results from the polynomial regression task
\vspace{-1.5cm}
}
\end{figure}

\clearpage
\end{appendices}